\documentclass[12pt,a4paper]{article}

\usepackage[papersize={8.5in,11in},left=1in, right=1in, top=1in, bottom=1in]{geometry}
\usepackage{amsmath, amssymb, amsthm,dsfont} 
\usepackage{graphicx,color}
\usepackage{array,eqnarray}
\usepackage{float}
\usepackage{multirow}
\usepackage{tikz}
\usetikzlibrary{shapes,arrows,backgrounds,fit,positioning}
\usepackage{arydshln}
\usepackage{lscape}
\usepackage[round]{natbib}
\usepackage{enumitem}
\usepackage{pifont}

\usepackage[footnotesize]{subfigure}
\usetikzlibrary{shapes,arrows}
\usepackage{array,booktabs,arydshln}
\usepackage{xcolor}

\newcommand{\comment}[1]{}

\usepackage[runin]{abstract}            
\abslabeldelim{\quad}						%
\newcommand{\HRule}{\rule{\linewidth}{0.5mm}}

\usepackage{lipsum}

\usepackage{algpseudocode}
\usepackage[ruled,vlined]{algorithm2e}
\makeatletter
\newcommand{\nosemic}{\renewcommand{\@endalgocfline}{\relax}}
\newcommand{\dosemic}{\renewcommand{\@endalgocfline}{\algocf@endline}}
\makeatother

\let\oldnl\nl
\newcommand{\nonl}{\renewcommand{\nl}{\let\nl\oldnl}}

\DeclareMathOperator{\bx}{\mathbf{x}}
\DeclareMathOperator{\bz}{\mathbf{z}}
\DeclareMathOperator{\by}{\mathbf{y}}

\DeclareMathOperator{\equals}{\enskip=\enskip}

\DeclareMathOperator{\cala}{\mathcal{A}}

\DeclareMathOperator{\calu}{\mathcal{U}}

\theoremstyle{plain}

\theoremstyle{definition}

\theoremstyle{theorem}

\theoremstyle{plain}

\theoremstyle{plain}

\theoremstyle{plain}

\theoremstyle{definition}

\begin{document}
\begin{center}
\large \textbf{
Injecting Author Information Into Word Embedding Space} \\
{\small \textit{The Knowledge Lab}}\\\vspace{2mm}
{\small Jamshid Sourati {\tt (jsourati@uchicago.edu)}}
\end{center}\vspace{-0.5cm}
\HRule

Here, we ask the question of how to incorporate information from the author space to the patterns captured by an word embedding model that is obtained by an embedding model such as word2vec. The incorporation could be performed in several ways:
\begin{enumerate}
    \item One of the most basic ideas is to represent authors by the set of words that they have used in their past publications (in a bag-of-words fashion). The authors could then be directly related to the embedding space based on the bags. However, one downside of this approach is that it is not sophisticated enough to capture hidden high dimensionality intra-relationships among the authors and inter-relationships between authors and words.
    \item Defining an author embedding space independently from the word embedding space. The main question is how to relate these two spaces using an intermediary space. An alternative but similar approach is to define an embedding for the author views.
    \item A more moderate method for incorporating author-based knowledge without explicitly embedding them into a new space is to build a graph on top of papers, where the nodes will be the words and the edges will be the papers (or alternatively, the authors). If we take papers to be linkages between the authors and words, it is more natural to use hypergraphs rather than regular graphs, since each edge (paper) can connect multiple authors with multiple words (concepts).
\end{enumerate}

\section{Using Hypergraphs to Model Author-Concepts Relationships}
Suppose $G = (V,E)$ is a hypergraph with nodes $V=A\cup W$, where $A$ and $W$ are the set of all authors and words (concepts), respectively,  and $E$ is that set of hyperedges. Each hyperedge $e\in E$ is as unordered subset of nodes ($e\subset V, \forall e\in E$). Such complex structure provides us with a tractable way of computing how accessible two concepts are given the set of authors (articles) and the concepts they have used in the past. This accessibility can be measured via transition probability of an unbiased random walk defined over the nodes $V$. The transition from one concept $w_1\in W$ to another $w_2\in W$ could occur through one or several authors. For now, we consider transitions of length two (via one author: $w_1\to a_1 \to w_2$) and length three (via two authors: $w_1 \to a_1\to a_2\to w_2$).

\subsection{Transitions with Length two}
This case has been already analyzed by~\citet{shi2019science}. We bring the details here in order to be self-contained and also to make it easier to analyze the next case. Before starting the analysis, let us define few more notations. We use  $d(v)$ to denote degree of node $v$, which is defined as the number of edges that contain $v$, i.e. $d(v)=|\{e\in E| v\in e\}|$. Size of a hyperedge is defined as the number of distinct nodes in it, i.e. $d(e)=|e|$. We also define the neighborhood set $\Gamma(v)$ to be the set of all nodes that share at least one hyperedge with $v$, i.e. $\Gamma(v)=\{u\in V|\exists e\in E: \{v,u\}\subset e\}$. We also define author-restricted neighbors of $v$ to be $\Gamma_A(v) = \Gamma(v)\cap A$.

We denote a specific random walk of length $t$ on the hypergraph by a sequence of nodes $\{n_0,n_1,n_2,...,n_t\}, n_i\in V, \forall i$. Starting from conceptual node $w_1\in W$, a single step in a hypergraph random walk has the following two parts (based on Model 2 suggested in \citet{cooper2011cover}):
\begin{enumerate}
    \item Randomly choosing a hyperedge that includes $w_1$
    \item Randomly selecting one of the nodes inside the chosen hyperedge (except $w_1$)
\end{enumerate}

Therefore, the probability of transitioning from $w_1\in W$ to $a\in A$ is
\begin{equation}
    \label{eq:trans_prob}
    \mathbb{P}(n_1=a|n_0=w_1) \equals \frac{1}{d(w_1)}\sum_{e: w_1\in e}\frac{\mathds{1}(a\in e)}{d(e)} \equals \frac{1}{d(w_1)}\sum_{e: \{a,w_1\}\subset e}\frac{1}{d(e)},
\end{equation}
where $\mathds{1}()$ is an indicator function.
Note that if $a\notin \Gamma(w_1)$ the probability of this single-step transition be zero. Similarly, the probability of the transition from word $w_1\in W$ to another word $w_2\in W$ via the author node $a\in A$ can be written by breaking the walk into two individual, independent steps\footnote{A random walk naturally defines a Markov chain and, therefore, each step only depends on the origin node and the destination independent of all the previous steps.}:
\begin{align}
    \label{eq:two_step_prob}
    \mathbb{P}(n_2=w_2, n_1=a | n_0=w_1) &\equals \mathbb{P}(n_1=a | n_0=w_1)\mathbb{P}(n_2=w_2|n_1=a) \\
    &\equals \left(\frac{1}{d(w_1)}\sum_{e: w_1\in e}\frac{\mathds{1}(a\in e)}{d(e)}\right)\left(\frac{1}{d(a)}\sum_{e': a\in e'}\frac{\mathds{1}(w_2\in e')}{d(e')}\right) \nonumber \\
    &\equals \frac{1}{d(w_1)}\frac{1}{d(a)} \left(\sum_{e: \{w_1,a\}\subset e}\frac{1}{d(e)}\right)\left(\sum_{e': \{a,w_2\}\subset e'}\frac{1}{d(e')}\right) \nonumber 
\end{align}
\textbf{}From~(\ref{eq:two_step_prob}), we can see that $a$ should be neighbor to both word end-points in order to get a non-zero transition probability for this path. That is, if $a\notin \Gamma(w_1)\cup\Gamma(w_2)$, then this probability will be zero. This implies that a length-two transition from $w_1$ to $w_2$ is only possibly through authors who have already published on both concepts. 

From this, we can write the probability of a length-two transition from $n_0=w_1$ to $n_2=w_2$ via any author node by summing up over all possible authors in the path:
\begin{align}
    \mathbb{P}(n_2=w_2, n_1\in A|n_0=w_1) \equals \sum_{a\in\Gamma_A(w_1)\cap\Gamma_A(w_2)} \mathbb{P}(n_2=w_2, n_1=a | n_0=w_1)
\end{align}

Now, we can compute transition probability of length-two path between $w_1$ and $w_2$ via an author node as the average of transition probabilities from one end-point to the other, i.e.  $\frac{1}{2}\big[\mathbb{P}(n_2=w_2, n_1\in A|n_0=w_1) + \mathbb{P}(n_2=w_1, n_1\in A|n_0=w_2)\big]$ corresponding to paths $w_1\to n_1\to w_2$ and $w_2\to n_1\to w_1$, respectively, which can be expanded as below
\begin{align}
    &\frac{1}{2} \sum_{a\in\Gamma_A(w_1)\cap\Gamma_A(w_2)} \frac{1}{d(w_1)}\frac{1}{d(a)} \left(\sum_{e: \{w_1,a\}\subset e}\frac{1}{d(e)}\right)\left(\sum_{e': \{a,w_2\}\subset e'}\frac{1}{d(e')}\right) \enskip + \nonumber\\
    &\frac{1}{d(w_2)}\frac{1}{d(a)} \left(\sum_{e: \{w_2,a\}\subset e}\frac{1}{d(e)}\right)\left(\sum_{e': \{a,w_1\}\subset e'}\frac{1}{d(e')}\right) \\
    =\enskip &\frac{1}{2}\left(\frac{1}{d(w_1)} + \frac{1}{d(w_2)}\right)\sum_{a\in\Gamma_A(w_1)\cap\Gamma_A(w_2)} \frac{1}{d(a)} \left(\sum_{e: \{w_1,a\}\subset e}\frac{1}{d(e)}\right)\left(\sum_{e': \{a,w_2\}\subset e'}\frac{1}{d(e')}\right) \nonumber
\end{align}

\subsection{Transitions with Length Three}
Transition probability from $w_1$ to $w_2$ with length three can be written based on the derivation from length-two transition. Consider the path $w_1=n_0\to n_1\to n_2\to n_3=w_2$ such that $n_1,n_2\in A$, then the transition probability $\mathbb{P}(n_3=w_2,n_2\in A, n_1\in A|n_0=w_1)$ can be written using the Markovian property of random walks
\begin{align}
    \mathbb{P}(n_1\in A|n_0=w_1)\cdot\mathbb{P}(n_2\in A|n_1\in A)\cdot\mathbb{P}(n_3=w_2|n_2\in A)
\end{align}
For fixed author nodes $n_1=a$ and $n_2=a'$, we will have
\begin{align}
    \label{eq:three_step_prob}
    &\underbrace{\mathbb{P}(n_1=a|n_0=w_1)\cdot\mathbb{P}(n_2=a'|n_1=a)}\cdot\underbrace{\mathbb{P}(n_3=w_2|n_2=a')} \equals \\\nonumber
    &\left[ \frac{1}{d(w_1)}\frac{1}{d(a)} \left(\sum_{e: \{w_1,a\}\subset e}\frac{1}{d(e)}\right)\left(\sum_{e': \{a,a'\}\subset e'}\frac{1}{d(e')}\right)\right]\cdot\left[\frac{1}{d(a')}\sum_{e'':\{a',w_2\}\in e''}\frac{1}{d(e'')}\right]
\end{align}
where the first two terms are expanded according to~(\ref{eq:two_step_prob}). In order to consider all possible author nodes for $n_1$ and $n_2$, we can either sum up the terms in~(\ref{eq:three_step_prob}) for all $a\in\Gamma_A(w_1)$ and $a'\in\Gamma_A(w_2)$. However, we can further restrict these sets by either limiting $a$ to $\Gamma_A(w_1)\cap\Gamma_A(a')$ or $a'$ to $\Gamma_A(a)\cap\Gamma_A(w_2)$. We chose the first option to write:
\begin{align}
    \frac{1}{d(w_1)}\sum_{\substack{a'\in\Gamma_A(w_2)\\a\in \Gamma_A(a')\cap\Gamma_A(w_1)}}\frac{1}{d(a)} \left(\sum_{e: \{w_1,a\}\subset e}\frac{1}{d(e)}\right)\left(\sum_{e': \{a,a'\}\subset e'}\frac{1}{d(e')}\right)\left[\frac{1}{d(a')}\sum_{e'':\{a',w_2\}\in e''}\frac{1}{d(e'')}\right]
\end{align}

\subsection{Matrix Manipulations}
It is much more efficient to compute the transition probabilities of single- or multi-step paths with matrix operations. We will compute a transition probability matrix $P$ ($|V|\times|V|$) whose $(i,j)$-th element indicates probability of transitioning from node $v_i$ to node $v_j$, i.e. $P_{ij} = \mathbb{P}(n_t=v_j|n_{t-1}=v_i)$. Each row and column is associated with a node in the hypergraph, which can be author or conceptual nodes. Note that each row of $P$ forms a probability mass function (PMF) corresponding to the probability of moving to other nodes from node $i$. However, the columns do not necessarily sum up to one. 

The transition probability can be computed using vertex weight matrix $R$, a $|E|\times|V|$ whose $(i,j)$-th element is one if node $v_i$ is in edge $e_j$ and zero otherwise. By definition, sum of elements in the  $i$-th row of $R$ equals to $d(e_i)$ and sum of elements in the $j$-th column is $d(v_j)$. Based on the equation~(\ref{eq:trans_prob}), probability matrix can be computed as
\begin{equation}
    \label{eq:matrix_trans_prob}
    P \equals D_E^{-1}RD_V^{-1}R^\top,
\end{equation}
where $D_E$ and $D_V$ are diagnal matrices containing edge degrees and node degrees, respectively.

Probability of multi-step transitions can be computed easily. For example, for two steps transitions $v_i\to v_j\to v_k$, we simply have $P(i,j)P(j,k)$. Considering all possible two-step transitions from $w_1$ to $w_2$ through an author node, the underlying probability will be
\begin{align}
    \mathbb{P}(n_2=w_2,n_1\in A|n_0=w_0) &\equals \sum_{a\in A} P(w_1,a)P(a,w_2) \nonumber\\
    &\equals P(w_1,[A])\cdot P([A],w_2),
\end{align}
where $P(w_1,[A])$ is the $w_1$-th row whose elements are associated with author columns (nodes in $A$). Similar notation is used for $P([A],w_2)$. More generally, the probability of $t$-step transition from $w_1$ to $w_2$ through $t-1$ author nodes ($t\geq2$) is $P(w_1,[A])\cdot P([A],[A])^{t-2} P([A],w_2)$.

\section{Word Embedding and Social Structure}
One important question is how to combine the structural information from the hypergraph network explained above and the semantic information provided by the word embedding. 

The idea of generating samples from the hypergraph in short, truncated random walks can be inspired by DeepWalk algorithm~\citep{perozzi2014deepwalk}. DeepWalk mimics natural language modeling on graphs by sampling training data from the network via random walks starting from various nodes. These truncated random walks are a sequence of nodes and therefore will be treated just like sentences, which could be windowed and paired to maximize the co-occurrence likelihood probabilities. Naturally, these randomly generated sequences can be combined with the real sentences when training (or fine-tuning) the word embedding model. On the other hand, node2vec algorithm~\citep{grover2016node2vec} suggested to use a biased random walk instead. Specifically, in order to keep a balance between local and exploratory search strategies such as Breadth First Search (BFS) and Depth First Search (DFS), respectively, they introduced a new random walk as a second-order Markovian chain. Each new move in their random walk depends on both the current and the previous node. Their sampling strategy had two tunable parameters (return and in-out parameters) that they trained using relatively small labeled data set.

\subsection{Combining Deepwalk and Text Embeddings}
We are interested in combining these two embeddings in a bidirectional way. That is, the model should give us the ability to take deepwalk-based similarities between entities and property nodes into account negatively. One idea is to simply me able toix and shuffle the sentences and then train a word2vec model. But this does not efficiently combine the information as the number of sentences are not balanced. Moreover, it is uni-directional only considering positive combination.

A second idea, is to combine the loss functions. For a given (word,context) pair with indices $(i,j)$ the original negative sampling loss for word2vec is \emph{(formal definition of notations is skipped for now)}:
\begin{align}
    \#(i,j)\log\sigma\left(\bx_i^\top\by_j \right) \enskip+\enskip k\cdot\mathbb{E}_{t\sim\mathbb{P}_n}\left[\log \sigma\left(-\bx_i^\top\by_t\right)\right],
\end{align}
minimizing which can \emph{almost} also be formulated as a matrix factorization problem. That is, in the solution the inner product of the embedding and context vectors (input and output layer weights) will have the form
\begin{equation}
    \bx_i^\top\by_j \equals \frac{|V|\cdot\#(i,j)}{k\cdot\#(i)\#(j)},
\end{equation}
where $V=\{w_1,...,w_n\}$ is the vocabulary and $\#(i)=\sum_j\#(i,j)$. The similarities between the words can be simply taken as the inner-product of the \emph{shifted positive} PMI (SPPMI) matrix $\mathbf{M}$, whose $(i,j)$-th element will be logarithm of $\bx_i^\top\by_j$ shown above minus $\log k$, restricted to positive values only. For instance, the similarity between the property node and the $i$-th entity will be the inner-product of the following two vectors:
\begin{align*}
    \mathbf{M}_{\mbox{\tiny kw}} &\equals \left[\bx_{\mbox{\tiny kw}}^\top\by_1 ,...,\bx_{\mbox{\tiny kw}}^\top\by_i, ...  \bx_{\mbox{\tiny kw}}^\top\by_n\right] \\
    \mathbf{M}_i &\equals \left[\bx_i^\top\by_1 ,..., \bx_i^\top\by_{\mbox{\tiny kw}}, ...  \bx_i^\top\by_n\right]
\end{align*}

Another way of computing the similarities is to use the SVD factorization matrices (normalized $\mathbf{U}$ or $\mathbf{V}$ matrices). 

The original loss function shown above can be combined with deepwalk loss in the following way:
\begin{align}
    &\frac{1}{|V|}\sum_{(i,j)}\left\{\#(i,j)\log\sigma\left(\bx_i^\top\by_j \right) \enskip+\enskip k\cdot\mathbb{E}_{t\sim\mathbb{P}_n}\left[\log\ \sigma\left(-\bx_i^\top\by_t\right)\right]\right\} \nonumber\\
    +\enskip & \frac{\alpha}{|V_{\mbox{\tiny dw}}|}\sum_{(i',j')} \#(i,j)\log\sigma\left(\bx_{i'}^\top\by_{j'} \right)
\end{align}
where $\alpha$ can be negative as well. This boils down to having a modified SPPMI matrix, where th pre-logarithm PMI values (association scores) would be modified to
\begin{align}
    \frac{|V|\cdot\#(i,j)}{k\cdot\#(i)\#(j)} \enskip+\enskip \frac{\alpha|V|^2\cdot\#_{\mbox{\tiny dw}}(i,j)}{k|V_{\mbox{\tiny dw}}|\cdot\#(i)\#(j)}
\end{align}
Modified SPPMI matrix can then be formed and similarities be defined similar to above. This model can take into account negative combination, but not much effectively. Note that $V_{\mbox{\tiny kw}$ (vocabulary of deepwalk sentences) only contains entities and the property term, therefore the modified elements in the SPPMI matrix will be limited to few terms. In the example above, the row corresponding to the $i$-th entity will be modified in at most $|V_{\mbox{\tiny kw}|$, including $\bx_i^\top\by_{\mbox{\tiny kw}}$. Even if the $i$-th entity and the property term are very similar according to the deepwalk model, negative $\alpha$ does not guarantee that this similarity will be reversed to very dissimilar words, since these two words can simply stay similar through other 
terms that do not exist in the deepwalk vocabulary (hence are not modified).

\begin{figure}[t]
    \centering
    \includegraphics[width=\textwidth]{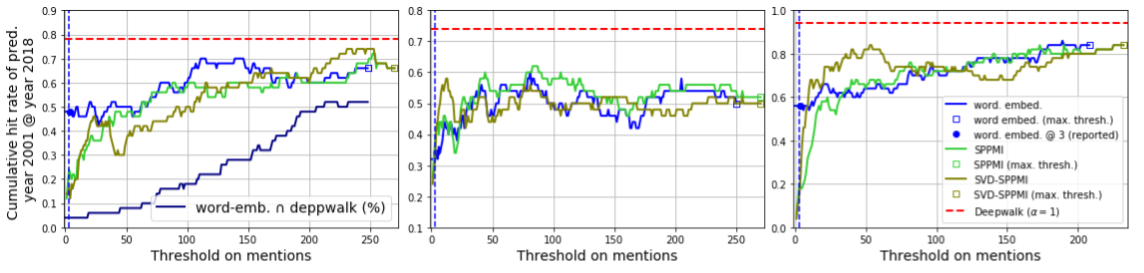}
    \caption{Hit rate of the predictions made by different variants of the word-embedding model with different thresholds over mention counts of the entities. The hit rate shown here is the cumulative hit rate of the predictions made at year 2001 computed at year 2018.}
\end{figure}

\bibliographystyle{apalike}
\bibliography{refs}
\end{document}


\includepdf[pages=-,pagecommand={},width=1.2\textwidth]{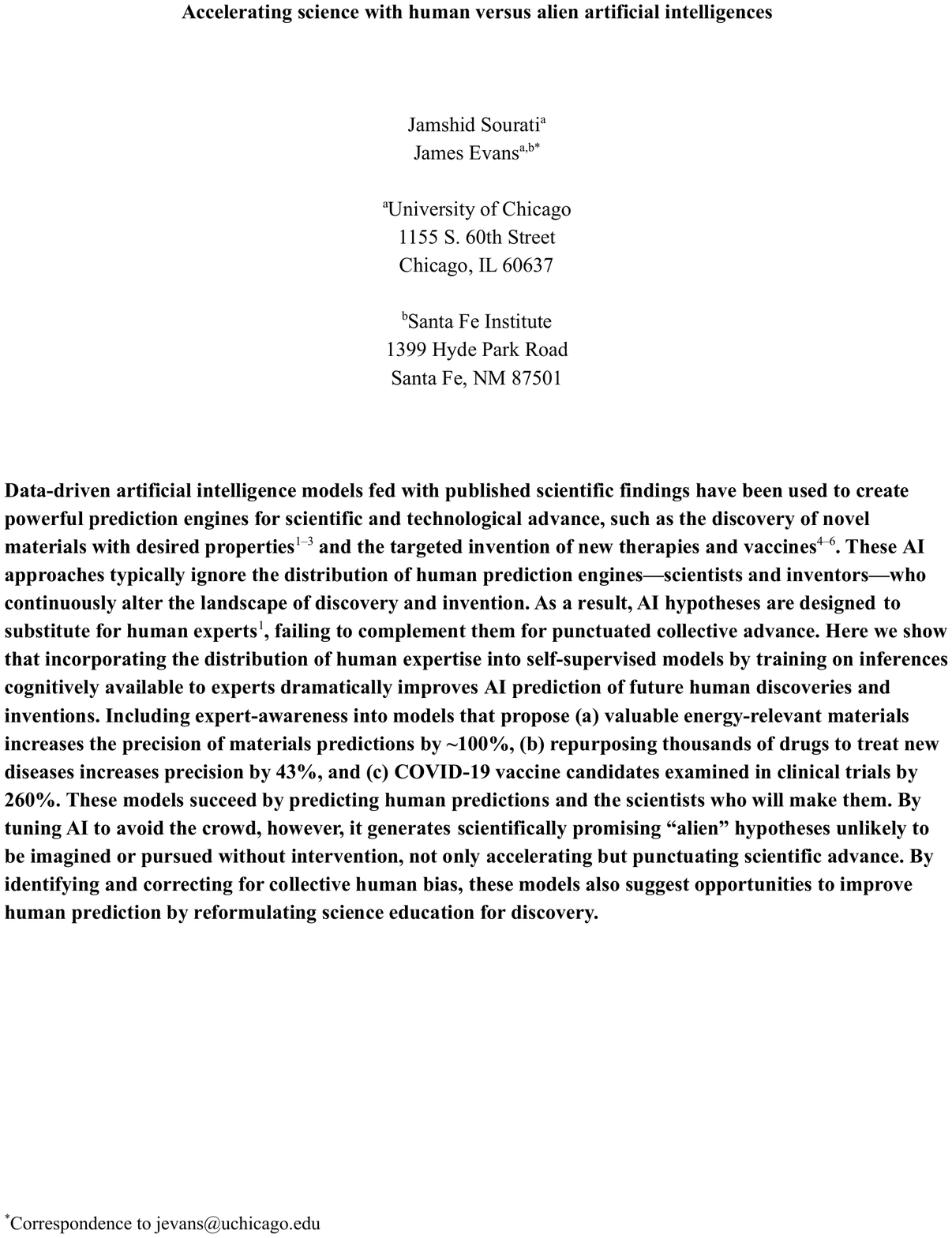}

\begin{center}
\Large \textbf{SUPPLEMENTARY INFORMATION: \\
ACCELERATING SCIENCE WITH HUMAN VS. ALIEN ARTIFICIAL INTELLIGENCES} \\[.5cm]
{Jamshid Sourati$^1$, James A. Evans$^{1,2}$}\\\vspace{.5cm}
{\small $^1$Knowledge Lab, University of Chicago, Chicago, IL, USA}\\
{\small $^2$Santa Fe Institute, Santa Fe, NM, USA}
\end{center}\vspace{-0.5cm}

\section{Multistep Transition Probabilities}
The first similarity metric we used based on our random walk settings was based on multistep transitions from the property node (denoted by $P$) to a target material (denoted by $M$). We considered two- and three-step transitions with intermediate nodes conditioned to belong to the set of authoring experts (denoted by $\cala$). In each case, the starting node $n_0$ is set to the property node and we compute the probability that a random walker reaches $M$ in two or three steps, i.e., $n_2=M$ or $n_3=M$, respectively. Therefore, the probability of a two-step transition through an intermediate author node is computed:
\begin{align}
    \mathbb{P}\big(n_2=M,n_1\in\cala\big|n_0=P\big) &\equals \sum_{A\in\cala}\mathbb{P}\big(n_2=M, n_1=A\big|n_0=P\big)\nonumber\\
    &\equals \sum_{A\in\cala}\mathbb{P}\big(n_1=A\big|n_0=P\big)\cdot\mathbb{P}\big(n_2=M\big|n_0=A\big),
\end{align}
where the second line draws on the independence assumptions implied by the Markovian process of random walks. Similar formulation could be derived for three-step transition. The individual transition probabilities in the second line are readily available based on our definition of a hypergraph random walk. For example, for a classic random walk with uniform sampling distribution, we get
\begin{equation}
    \mathbb{P}\big(n_1=A\big|n_0=P\big)\equals \frac{1}{d(P)}\sum_{e:\{P,A\}\in e}\frac{1}{d(e)},
    \label{eq:individual_transition_prob}
\end{equation}
where $d(P)$ is the degree of node $P$, i.e., the number of hyperedges it belongs to, and $d(e)$ is the size of hyperedge $e$, i.e., the number of distinct nodes inside it. The first multiplicand in the right-hand side of~(\ref{eq:individual_transition_prob}) accounts for selecting a hyperedge that includes $P$ and the second computes the probability of selecting $A$ from one of the common hyperedeges (if any). 

The above computations can be compactly represented and efficiently implemented through matrix multiplication. Let $\mathbf{P}$ represent the transition probability matrix over all nodes such that $\mathbf{P}_{ij}=\mathbb{P}\big(n_1=j\big|n_0=i\big)$. Then, two- and three-step transitions between nodes $P$ and $M$ could be computed via $\mathbf{P}\big(P,[\cala]\big)\cdot\mathbf{P}\big([\cala],M\big)$ and $\mathbf{P}\big(P,[\cala]\big)\cdot\mathbf{P}\big([\cala],[\cala]\big)\cdot\mathbf{P}\big([\cala],M\big)$, respectively, where $\mathbf{P}\big(P,[\cala]\big)$ defines selection of the row corresponding to node $P$ and columns corresponding to authors in set $\cala$.

\section{Combining Scores for Human (Un)availability and Scientific Plausibility}
Our Alternative or Alien Artificial Intelligence (AAI) algorithm combines two sources of information to generate candidates that are  simultaneously alienated and scientifically plausible. The two signals we use to quantify these components in our experiments include the Shortest-Path distance (SP-$d$) of the materials to the property node (measuring human expert avoidance or cognitive unavailability) and their semantic similarities based on a word embedding model with regards to the property keyword (measuring relevance). Following \cite{tshitoyan2019unsupervised}, the word embedding model we used was the skipgram Word2Vec model trained over the literature in the five-year period preceding the prediction year. The mixing coefficient $\beta\in[0,1]$ determines how much importance will be assigned to avoiding authors versus chasing current theoretical plausibility when combining scores. It is also desirable that the effect of the two sources become equal when $\beta=\frac{1}{2}$, and that the output score varies continuously as $\beta$ changes. Let us denote human avoidance and scientific plausibility scores computed for an entity $x$ by $s_1(x)$ and $s_2(x)$, respectively. In this section, we discuss several methods for combining these signals and compare them in the context of AAI's performance.

Simply weighted averaging of the scores through $\beta s_1(x)+(1-\beta)s_2(x)$ is inappropriate for our experiment due to highly distinct scales of the human avoidance and scientific plausibility signals (preventing equal contribution when $\beta=1/2$). Moreover, the SP-$d$ values are unbounded as they can become arbitrarily large for entities disconnected from the property node in our hypergraph. As a result, Z-scores could not be directly applied. As a workaround, we applied Van der Waerden transformation over the scores. Suppose $S$ is a set of scores and $s(x)\in S$, then its Van der Waerden transformation $\tilde{s}(x)$ is defined as
\begin{equation}
    \tilde{s}(x) \equals \phi\left(\frac{r(x)}{|S|+1}\right),
\end{equation}
where $\phi$ is the quantile function of the normal distribution, $r(x)$ is the rank of $s(x)$ within the set $S$ and $|S|$ denotes the cardinality of $S$. We will then take the weighted average of Z-scores for the transformed signals $\tilde{s}_1(x)$ and $\tilde{s}_2(x)$ for each material $x$ as the ultimate hybrid score to be used in our final ranking. 

Alternatively, geometric and harmonic means could be used for combining variables with unequal scales. In order to be able to use these measures, we replaced unbounded values of SP-$d$ by an arbitrary finite value whose magnitude is larger than the other finite elements. We define the $\beta$-weighted versions of these means as below:
\begin{align}
    \mbox{geometric:}\quad &s_{\mbox{\tiny GEO}}(x)=\bigg(s_1(x)^\beta\cdot s_2(x)^{1-\beta}\bigg)^{1/2} \\[.25cm]
    \mbox{harmonic:}\quad &s_{\mbox{\tiny HRM}}(x)=2\bigg/ \left(\frac{\beta}{s_1(x)} + \frac{1-\beta}{s_2(x)}\right)
\end{align}

In order to make comparison between the above $\beta$-weighted combination techniques in the context of chemical compounds and the thermoelectric property, we set $s_1$ and $s_2$ to SP-$d$ and Power Factor (PF) values, respectively. We ran AAI experiments varying $\beta$ values between 0 and 1 with steps of 0.1 and used the same metrics to self-evaluate the human avoidance and scientific plausibility of candidates generated with each $\beta$ value. Results indicate how rapidly changing the mixture weight shifts the attention from scientific relevance to human unavailability in various formulations. These results do not evaluate the performance of the AAI framework, but merely act as sanity checks for the appropriateness of score combination methods.

\begin{figure}
    \centering
    \input{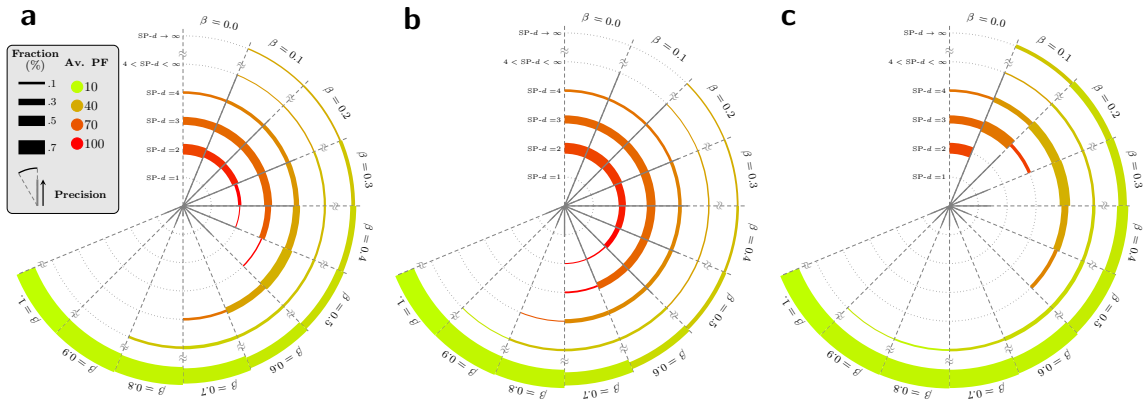}
    \caption{Distribution of SP-$d$ values (alienness) and PF values (scientific relevance) of candidates generated by our AAI framework using various $\beta$-weighted score combinations: \textbf{(a)} Van der Waerden transformation followed by weighted averaging of the Z-scores, \textbf{(b)} weighted geometric mean, and \textbf{(c)} weighted harmonic mean.}
    \label{fig:combination_methods_aai}
\end{figure}

Results shown in fig.~\ref{fig:combination_methods_aai} demonstrate that using the harmonic mean leads to abrupt drops in SP-$d$ values when $\beta$ slowly departs from zero. By contrast, using the geometric mean causes oversmoothing and yields asymmetric behaviour around $\beta=0.5$ overrating the scientific relevance. Applying Z-scores of the transformed scores resulted in a reasonably balanced and smooth $\beta$-weighted combination of scores.

\section{Experiments with Deep Neural Networks}
We evaluated the effect of incorporating distribution of expertise in our predictive models after replacing our deepwalk method with deeper graph convolutional neural networks. 

\subsection{Graph Neural Networks}

Graph Neural Networks (GNNs) have become a popular tool for learning low-dimensional graph representations or solving high-level tasks such as classification of graph nodes~\cite{chami2021machine}. They owe this popularity to their unique and efficient way of exploiting graph connectivities to propagate information between a central node and its neighborhood, their ability to incorporate feature vectors for nodes and/or edges, and their superior generalization to unseen (sub)graphs. Similar to deepwalk, these models build a low-dimensional embedding space where graph-based similarities are preserved. However, unlike deepwalk, they incorporate node feature vectors and directly utilize graph connectivities for message passing between nearby nodes when constructing the embedding space. 

The embedding vector of a central node is constructed by sequentially processing messages passed from its local neighbors. There are numerous ways of aggregating the signals reaching out from neighbors. In our experiments, we used the Graph Sample and Aggregate (GraphSAGE) platform, which applies the aggregations function on a subset of neighbors to avoid computational overhead~\cite{hamilton2017inductive}. Let $\bh_i^\ell$ denote the message from the $i$-th node in the $\ell$-th step of this sequential procedure. Then, the representation of the $i$-th node at the next level will be computed as
\begin{equation}
    \bh_i^{\ell+1} \equals \sigma\left( \mathbf{f}_{\mbox{\tiny AGG}}\left(\{\bh_j\}_{j\in \tilde{\caln}(i)}\right) \mathbf{W}_\ell \right),
\end{equation}
where $\mathbf{f}_{\mbox{\tiny AGG}}$ is an aggregation function (e.g., mean, pooling, etc) applied on the concatenation of the local neighborhood's messages $\{\bh_j\}_{j\in \tilde{\caln}(i)}$, where $\tilde{\caln}(i)$ is a subset of $k_\ell$ uniformly sampled nodes from the immediate neighbors of the $i$-th node $\caln(i)$. The resulting aggregated messages will undergo a single-layer neural network parameterized by $\mathbf{W}_\ell$ (the bias term is ignored for simplicity) and the non-linear activation $\sigma$. The input messages in the first step, i.e., $\bh_i^0 \forall i$, are set to the provided node feature vectors. The final representation of the $i$-th node will be reached after $L$ steps. We used the same set of hyperparameters as the original paper~\cite{hamilton2017inductive}; we considered two steps ($L=2$) with samples sizes $k_1=25$ and $k_2=10$. We also used the mean aggregation function and applied the non-linearity through Rectified Linear Unit (ReLU) activation. 

We approached the discovery prediction problem in an unsupervised manner through a graph autoencoder~\cite{kipf2016variational}, where the encoder component was modeled using the GraphSAGE architecture and the decoder component simply consisted of a parameter-less inner-product of the encoder's output. This autoencoder was trained by minimizing a link-prediction loss function, which was approximated with negative sampling. The approximate loss has two parts accounting for the similarity of positive samples (pairs of nearby nodes) and the dissimilarity of negative samples (pairs of unconnected nodes). 


Our mechanism of sampling positive and negative pairs closely resembled that which was used in deepwalk: the former is formed by pairing central/contextual nodes within windows sliding over short random walks, and the latter by means of sampling from the unigram distribution raised to power $3/4$ over the full set of nodes~\cite{mikolov2013distributed}. Once the positive samples were drawn using sliding window size of 8, we begin minimizing the loss function in a mini-batch setting by iterating over pairs. We used batch size of 1000, negative sampling size of 15 (per positive pair), learning rate of $5\times10^{-6}$ and the Adam optimizer~\cite{kingma2014adam} with the default parameters. 

\subsection{Experimental Settings}

We trained our graph autoencoder in two different settings: (1) using our full hypergraph, and (2) after dropping author nodes. In both settings, we only considered the material and property nodes. In the full setting, we took account of author nodes at the time we draw positive samples and compute the adjacency matrix. In this setting the positive samples were drawn from deepwalk sequences associated with $\alpha=1$, whereas the experiment without authors used sequences corresponding to $\alpha\to\infty$ so that no author nodes would be present in the random walk sequences. Moreover, connectivities between nodes were different for the two settings. In the author-less network, we connected two property or material nodes only if they appeared in the same paper. In the full setting, we kept these connections and added more edges between nodes with at least one common author neighbor (even in the absence of papers in which they co-occur).

\bibliography{refs}